\documentclass[9pt,conference]{IEEEtran}
\IEEEoverridecommandlockouts
\usepackage{amsmath,graphicx}
\usepackage{enumitem}
\usepackage{adjustbox}
\usepackage{amsmath,graphicx}
\usepackage{multirow}
\usepackage{marvosym}
\usepackage{multicol}
\usepackage{hyperref}
\usepackage{amssymb}
\usepackage{tabularx}
\usepackage{caption}
\usepackage{graphicx}
\usepackage{fontawesome}
\usepackage{booktabs}
\usepackage{cite}
\usepackage{amsmath,amssymb,amsfonts}
\usepackage{algorithmic}
\usepackage{graphicx}
\usepackage{textcomp}
\usepackage[table,xcdraw]{xcolor}

\def\BibTeX{{\rm B\kern-.05em{\sc i\kern-.025em b}\kern-.08em
    T\kern-.1667em\lower.7ex\hbox{E}\kern-.125emX}}
\begin{document}
\title{
Prompting DirectSAM for Semantic Contour Extraction in Remote Sensing Images
\thanks{*~Equal contributions, 
\Letter~Corresponding author\par
}
}

\author{\IEEEauthorblockN{Shiyu Miao$^{*1}$, Delong Chen$^{*2}$, Fan Liu$^{1}$\textsuperscript{\Letter}, Chuanyi Zhang$^{1}$, Yanhui Gu$^{3}$, Shengjie Guo$^{3}$, Jun Zhou$^{4}$}
\IEEEauthorblockA{$^{1}$College of Computer Science and Software Engineering, Hohai University, Nanjing, China \\
$^{2}$Hong Kong University of Science and Technology, Hong Kong SAR \\
$^{3}$School of Computer and Electronic Information, Nanjing Normal University, Nanjing, China \\
$^{4}$School of Information and Communication Technology, Griffith University, Queensland, Australia \\
E-mail: 2206010417@hhu.edu.cn, delong.chen@connect.ust.hk, fanliu@hhu.edu.cn, \\ 20231104@hhu.edu.cn, gu@njnu.edu.cn, gshengjies@gmail.com, jun.zhou@griffith.edu.au
}

}

\maketitle

\begin{abstract}


The Direct Segment Anything Model (DirectSAM) excels in class-agnostic contour extraction. In this paper, we explore its use by applying it to optical remote sensing imagery, where semantic contour extraction—such as identifying buildings, road networks, and coastlines-holds significant practical value. Those applications are currently handled via training specialized small models separately on small datasets in each domain. We introduce a foundation model derived from DirectSAM, termed DirectSAM-RS, which not only inherits the strong segmentation capability acquired from natural images, but also benefits from a large-scale dataset we created for remote sensing semantic contour extraction. This dataset comprises over 34k image-text-contour triplets, making it at least 30 times larger than individual dataset. DirectSAM-RS integrates a prompter module: a text encoder and cross-attention layers attached to the DirectSAM architecture, which allows flexible conditioning on target class labels or referring expressions. We evaluate the DirectSAM-RS in both zero-shot and fine-tuning setting, and demonstrate that it achieves state-of-the-art performance across several downstream benchmarks.

\end{abstract}


\section{Introduction}
\label{sec:intro}

Direct Segment Anything Model (DirectSAM)~\cite{subobject} is a powerful contour extraction model that enables efficient ``segment everything'', generating comprehensive panoptic subobject-level segmentations. Based on the SA-1B dataset comprising 1 billion masks across 11 million images, DirectSAM's large-scale pretraining process consumed 11k NVIDIA A100 GPU hours. This extensive pretraining endowed the model with rich knowledge of visual contours. However, DirectSAM is \textbf{non-interactive} and \textbf{class-agnostic}. These constraints restrict DirectSAM's convenient use in applications beyond subobject-level image tokenization. Furthermore, since the model was pretrained primarily on natural images, its generalization to other domains, such as remote sensing imagery, remains limited.

In this paper, we introduce DirectSAM-RS, a \textit{vision-language foundation model} for semantic contour extraction in optical remote sensing imagery. As a foundation model~\cite{foundation,dalle2,blip,glip,bert,gpt3}, DirectSAM-RS offers both \textbf{high flexibility} and \textbf{strong performance}, mirroring advantages seen in foundation models from other domains such as CLIP~\cite{clip}, RemoteCLIP~\cite{remoteclip}, and Segment Anything Model (SAM)~\cite{SAM}. 

This foundation model represents a significant departure from existing models in semantic contour extraction for remote sensing. While this task has long been fundamental in remote sensing, attracting considerable attention, traditional approaches~\cite{dense,RCF,cat,ldc,weakly} have focused on training models separately for different targets. Such a methodology has limited model performance, particularly given the small scale of existing datasets~\cite{LRSNY,slsd,bube} (typically fewer than 1k samples). 

In contrast, DirectSAM-RS enables knowledge sharing across diverse semantic targets through joint multi-task learning. It stands out as the first vision-language foundation model for contour extraction capable of accepting free-form textual prompts that specify the target, as opposed to existing visual-only models. Moreover, it is the first model able to perform zero-shot semantic contour extraction without using any training samples from downstream datasets. 
To create DirectSAM-RS, we introduced novel technical contributions in both data and modeling, briefly described as follows:

\begin{itemize}
    \item \textbf{A Large-scale dataset for pretraining}. We constructed a semantic contour extraction dataset by repurposing existing semantic segmentation datasets with our proposed Mask2Contour (\texttt{M2C}) transformation. The \texttt{M2C} process produces a total of 34k image-text-contour triplets from LoveDA~\cite{loveda}, iSAID~\cite{isaid}, DeepGlobe~\cite{deepglobe}, and RefSegRS~\cite{rrsis} datasets. We name this resulting dataset \texttt{RemoteContour-34k}.

    \item \textbf{A prompter attached to DirectSAM}. We develop a prompter architecture, which consists of an encoder that extracts semantic information from the textual prompt, and cross-attention layers inserted into the DirectSAM. These cross-attention layers interleave with the image encoder blocks, facilitating multi-modal information fusion. This simple yet effective prompter design enables flexible conditioning of DirectSAM while preserving its pretrained knowledge.

\end{itemize}

We validate DirectSAM-RS on three downstream contour extraction datasets: SLSD~\cite{slsd} for coastline extraction, Beijing Urban Building Extraction (BUBE)~\cite{bube}, and LRSNY~\cite{LRSNY} for road extraction. We evaluated both zero-shot and fine-tuning modes. We observed that zero-shot DirectSAM-RS can already outperform some baseline methods, and when we fine-tuned the model, it established new state-of-the-art on all of three benchmarks. Specifically, DirectSAM-RS achieved ODS scores of 0.772, 0.887, and 0.958 on road, building, and coastline extraction benchmarks, respectively. These results represent significant relative performance improvements of 21\%, 5\%, and 7\% over previous best methods. To facilitate future research, the \texttt{RemoteContour-34k} dataset and codes will be fully open-sourced and are available at \url{https://github.com/StevenMsy/DirectSAM-RS}.

\section{Method}
\label{sec:pagestyle}

\begin{figure*}
\begin{center}
\includegraphics[width=\linewidth]{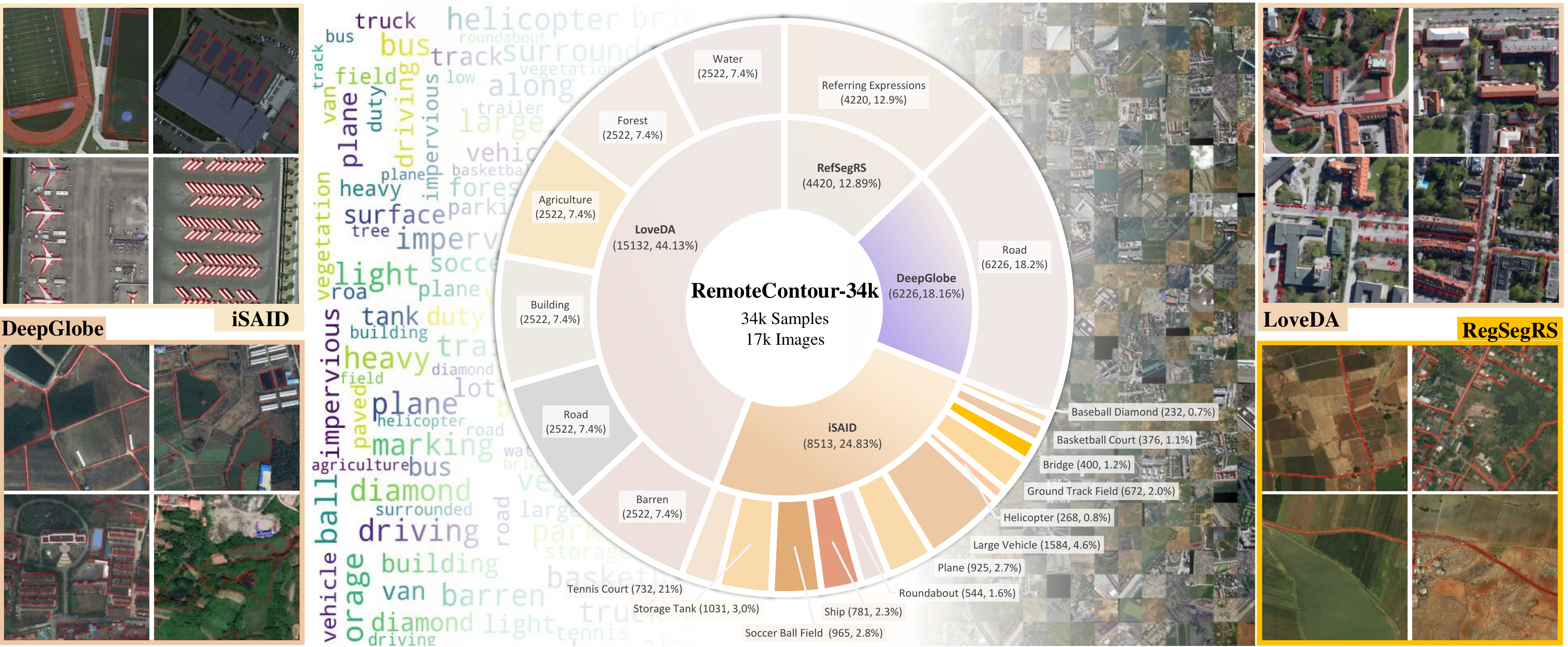}
\caption{\textbf{Composition of the proposed dataset.} We annotated the number of samples and the percentage of each subset and each class. The \texttt{RemoteContour-34k} dataset consists both rich semantics (as shown by the word cloud), and diverse visual domains (\textit{e.g.,} urban, rural).}

\label{fig:remotecontour}
\end{center}
\end{figure*}

\begin{figure}
\centering
\includegraphics[width=1\columnwidth]{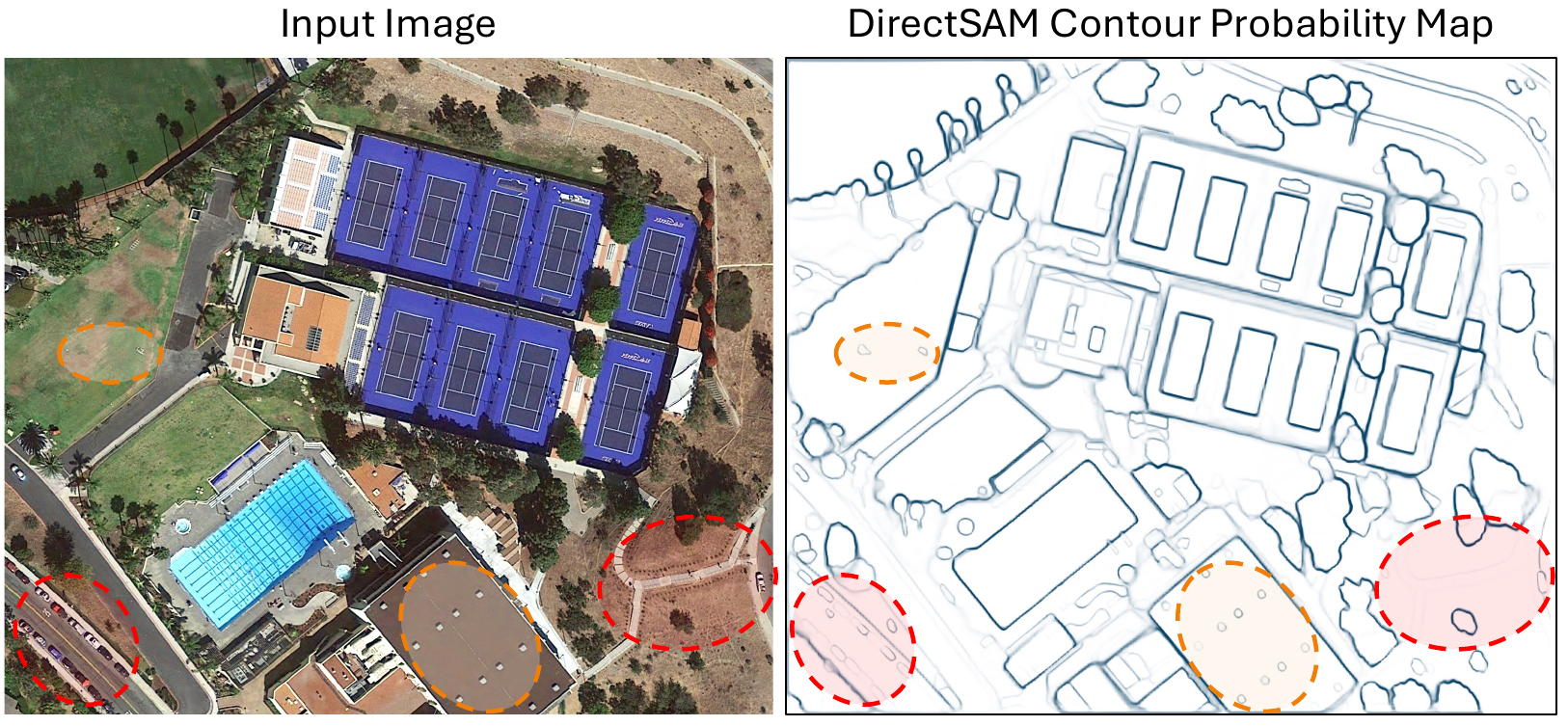}
\caption{While being able to successfully identify most key elements, the raw DirectSAM model suffers from \textcolor{red}{missing segmentation} of cars and roads, and \textcolor{orange}{over-segmentation} of part components. Moreover, the model extracts the contours of all semantic targets as it's class-agnostic. These issues limit its direct applicability for semantic contour extraction in remote sensing.}

\label{fig:DirectSAM}
\end{figure}

\subsection{Preliminary: DirectSAM}

Let's begin by briefly reviewing DirectSAM. It is a SegFormer-based model that takes an RGB image $I \in \mathbb{R}^{H \times W \times 3}$ as input and generates the corresponding contour probability map $\hat{Y} \in \mathbb{R}^{H \times W}$ in $(0,1)$. The model has acquired robust contour extraction capability through extensive pretraining on SA-1B, which is currently the largest image segmentation dataset and was also used to train the powerful SAM models. DirectSAM has demonstrated strong contour extraction capability on natural images, and here we show that it also has potential value for remote sensing. As exemplified in Fig.~\ref{fig:DirectSAM}, DirectSAM successfully extracted various visual elements in the given aerial images, such as buildings, tennis courts, and trees. However, several key limitations prohibit its direct application to remote sensing tasks, as discussed below.

Firstly, DirectSAM occasionally fails to detect certain important targets, such as cars and roads noted in Fig.~\ref{fig:DirectSAM}. This could be due to the partial label problem of SA-1B and the domain gap between SA-1B and remote sensing images. Second, DirectSAM was trained for subobject-level image segmentation, while semantic contour extraction in remote sensing usually requires the model to operate at the level of object entities. This over-segmentation of fragmented and excessively detailed part components will negatively impact the performance. Finally, DirectSAM is class-agnostic--it does not distinguish between different semantic categories when extracting contours. This limits its utility for remote sensing applications that often require class-specific structural analysis.

\begin{figure}[t]
\centering
\includegraphics[width=1\columnwidth]{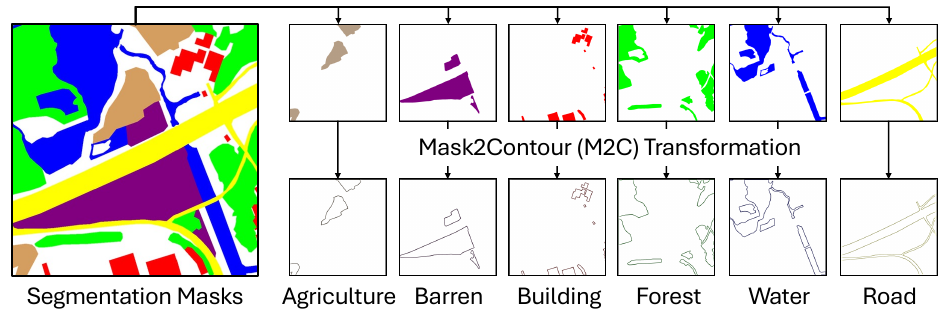}
\caption{Example of the proposed Mask2Contour (\texttt{M2C}) transformation. It enables us to repurpose the existing semantic segmentation dataset with mask annotations to the semantic contour extraction task.}
\label{fig:mask2contour}
\end{figure}

\begin{figure*}[t]
\begin{center}
\includegraphics[width=0.85\linewidth]{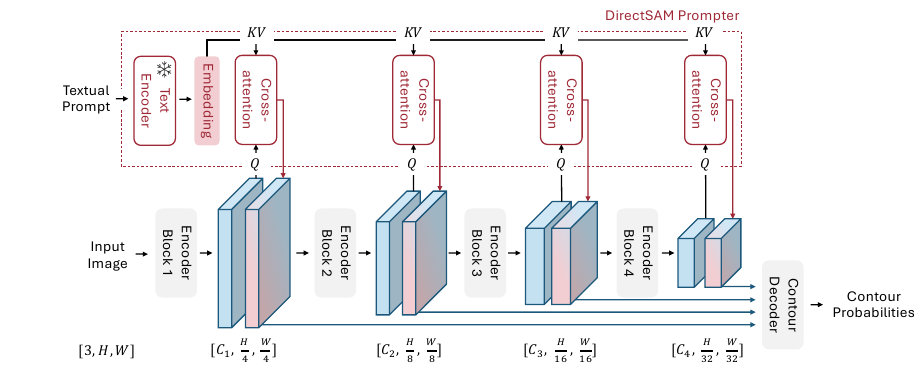}
\caption{\textbf{Model architecture of the proposed DirectSAM-RS}. We extend the base DirectSAM model (encoder blocks and contour decoder) with a prompter architecture (red). This prompter consists of a text encoder that extracts semantic information from textual prompts, and cross-attention layers that fuse the prompt information into visual feature maps at different stages.}

\label{fig:architecture}
\end{center}
\end{figure*}

\subsection{Overview of Our Methodology}

To address the above challenges, we propose DirectSAM-RS, a vision-language foundation model for semantic contour extraction in remote sensing. Our approach is motivated by the intuition that contour extraction capabilities for various targets in remote sensing share fundamental similarities. This insight suggests that a unified multi-tasked model, capable of extracting contours for multiple semantic targets, could outperform specialized models trained on individual tasks. By leveraging this shared knowledge across different tasks, we could create a more robust and generalizable model.

To realize this vision, our methodology incorporates two key elements: a large-scale, semantically diverse dataset spanning multiple contour extraction tasks, and a promptable architecture attached to DirectSAM that dynamically adapts the behavior based on the textual prompt. This combination allows the model to benefit from multi-task learning while maintaining the flexibility to address specific contour extraction tasks as needed. In the following sections, we will delve into the details of our dataset construction and model architecture.

\subsection{Dataset Construction}

To enhance DirectSAM-RS's textual comprehension and generalization capabilities for remote sensing imagery, we introduce the \texttt{RemoteContour-34k} dataset, a large-scale, diverse dataset tailored specifically for semantic contour extraction in remote sensing images. As illustrated in Fig.~\ref{fig:remotecontour}, \texttt{RemoteContour-34k} encompasses a substantial collection of 34,291 image-text-contour triplets with multiple distinct free-form textual queries. The dataset is curated from a variety of optical remote sensing sources, including semantic segmentation and instance segmentation datasets. This diverse sourcing ensures that \texttt{RemoteContour-34k} covers a broad spectrum of visual and textual contexts, providing a robust foundation for training the powerful DirectSAM-RS.

To maximize the dataset's efficacy, \texttt{RemoteContour-34k} incorporates samples from several key sources. Primarily, it integrates data from the RefSegRS dataset, complete with corresponding free-form textual queries. To further enhance visual diversity, samples from the LoveDA and iSAID segmentation datasets are included. Recognizing the significance of road-related features in remote sensing applications, we have also incorporated data from the DeepGlobe dataset. This addition addresses the need for a more comprehensive representation of road features, thereby improving the model's capacity to detect and interpret road elements in remote sensing imagery.

The M2C transformation is illustrated in Figure~\ref{fig:mask2contour}, where we utilize the border following algorithm\cite{findcontours} to transform segments to contours.


\subsection{Model Architecture}

As illustrated in Figure \ref{fig:architecture}, DirectSAM-RS adopted and enhanced the transformer-based visual encoder from DirectSAM, leveraging its ability to capture contour-related information from input images effectively. Additionally, we use a promptable text encoder to enable free-form query modeling.
Given an input image $ I \in \mathbb{R}^{H \times W \times 3}$ and query text $T$, a hierarchical visual DirectSAM (SegFormer) encoder is utilized to extract multiscale visual feature maps $ f_i \in \mathbb{R}^{C_i \times H_i \times W_i} $, where $ i \in \{1, 2, 3, 4\}$, and $ C_i$, $ H_i $, and $W_i$ represent the number of channels, height, and width of the feature maps from the $ i $-th stage, respectively. Meanwhile, the query $T$ is prompted as  \textit{"Edge of all \{$T$\}s "} and fed into the text encoder to obtain a features sequence $ E=\{e_{cls},e_1,e_2,...,e_L\}, E \in \mathbb{R}^{(L+1) \times D}$. Here, $L$ denotes the prompt sequence length and $D$ denotes the embedding dimension. 

The extracted visual feature $ f_i $ and text embedding $T$ will be fused through a cross-attention module to select the most important semantics from textual information. This process produces a set of fused features $\hat{f}_i$ that integrate both visual and textual information. Finally, an MLP-based lightweight contour decoder is utilized. This decoder consolidates all hierarchical multimodal feature maps and merges them into a unified feature map \( F \in \mathbb{R}^{\frac{H}{4} \times \frac{W}{4} \times 4C} \). Subsequently, another MLP layer processes this fused feature map to predict the final contour probability map~$ \hat{Y} \in \mathbb{R}^{H \times W}$.

\section{Experiments}
\label{sec:typestyle}

\subsection{Implementation Details}


In our study, we implemented the DirectSAM-RS using the Huggingface\cite{huggingface} framework based on PyTorch\cite{pytorch}. 
The backbone architecture of our DirectSAM-RS leveraged Segformer-b5, with its hierarchical feature channels set to \{64, 128, 320, 512\}.
Additionally, the DirectSAM Prompter incorporated cross-attention blocks, with the number of layers set to 2 and the tokens embedding dimension set to 512. DirectSAM-1800px-0424 weight was utilized as our initializing visual encoder parameter. BERT-base-uncased\cite{bert} pretrained model was utilized as the initial text encoder. During training and inference, all input images were resized to $1024 \times 1024$.

For optimization during training, we utilized the AdamW\cite{adamw} optimizer, setting $\beta_1 = 0.9$ and $\beta_2 = 0.95$, with a weight decay of 0.01. The training and validation processes were conducted on a Linux server equipped with 8 NVIDIA RTX 4090 GPUs, with a batch size of 64. During validation, to ensure a fair comparison, all previous methods and ours followed the same training and testing protocol. In the tests, the max-dist index $d_{max}$ for contour extraction was set to 0.0075 for all methods.

\subsection{Evaluation Metrics}

\begin{table}[]
\caption{Performance Comparison on 
LRSNY, BUBE and SLSD benchmark tasks. }
\label{tab:main}
\resizebox{\linewidth}{!}{%
\begin{tabular}{ccccc}
\toprule
\textbf{Dataset} &
  \textbf{Method} &
  \textbf{ODS} &
  \textbf{OIS} &
  \textbf{LineIoU@3} \\ \midrule
 &
  SwinT+OADecoder~\cite{oadecoder} &
  .494 &
  .822 &
  .270 \\
 &
  SegFormer-b5~\cite{xie2021segformer} &
  .635 &
  .884 &
  .391 \\
 &
  \cellcolor[HTML]{F3F3F3}DirectSAM~\cite{subobject}  \texttt{ZS} &
  \cellcolor[HTML]{F3F3F3}.202&
  \cellcolor[HTML]{F3F3F3}.634&
  \cellcolor[HTML]{F3F3F3}.065\\
 &
  \cellcolor[HTML]{F3F3F3}DirectSAM~\cite{subobject}  \texttt{FT} &
  \cellcolor[HTML]{F3F3F3}.650 &
  \cellcolor[HTML]{F3F3F3}.937 &
  \cellcolor[HTML]{F3F3F3}.395 \\
 &
  \cellcolor[HTML]{E2ECFD}DirectSAM-RS \texttt{ZS} &
  \cellcolor[HTML]{E2ECFD}.653 &
  \cellcolor[HTML]{E2ECFD}.911 &
  \cellcolor[HTML]{E2ECFD}.337 \\
\multirow{-6}{*}{\begin{tabular}[c]{@{}c@{}}LRSNY~\cite{LRSNY}\\ (Road)\end{tabular}} &
  \cellcolor[HTML]{E2ECFD}DirectSAM-RS \texttt{FT} &
  \cellcolor[HTML]{E2ECFD}\textbf{.772} &
  \cellcolor[HTML]{E2ECFD}\textbf{.962} &
  \cellcolor[HTML]{E2ECFD}\textbf{.455} \\ \hline
 &
 \rule{0pt}{2.2ex}
  BDCN~\cite{bdcn} &
  .536 &
  - &
  .385 \\
 &
  SDLED~\cite{bube} &
  .842 &
  - &
  .615 \\
 &
  SegFormer-b5~\cite{xie2021segformer} &
  .844 &
  .956 &
  .501 \\
 &
  \cellcolor[HTML]{F3F3F3}DirectSAM~\cite{subobject}  \texttt{ZS} &
  \cellcolor[HTML]{F3F3F3}.237 &
  \cellcolor[HTML]{F3F3F3}.555 &
  \cellcolor[HTML]{F3F3F3}.076\\
 &
  \cellcolor[HTML]{F3F3F3}DirectSAM~\cite{subobject}  \texttt{FT} &
  \cellcolor[HTML]{F3F3F3}.864 &
  \cellcolor[HTML]{F3F3F3}.971 &
  \cellcolor[HTML]{F3F3F3}.518 \\
 &
  \cellcolor[HTML]{E2ECFD}DirectSAM-RS \texttt{ZS} &
  \cellcolor[HTML]{E2ECFD}.705 &
  \cellcolor[HTML]{E2ECFD}.899 &
  \cellcolor[HTML]{E2ECFD}.329 \\
\multirow{-7}{*}{\begin{tabular}[c]{@{}c@{}}BUBE~\cite{bube}\\ (Building)\end{tabular}} &
  \cellcolor[HTML]{E2ECFD}DirectSAM-RS \texttt{FT} &
  \cellcolor[HTML]{E2ECFD}\textbf{.887} &
  \cellcolor[HTML]{E2ECFD}\textbf{.997} &
  \cellcolor[HTML]{E2ECFD}\textbf{.565} \\ \hline
 &
 \rule{0pt}{2.2ex}
  HED~\cite{HED} &
  .897 &
  .994 &
  .768 \\
 &
  SegFormer-b5~\cite{xie2021segformer} &
  .861 &
  .964 &
  .742 \\
 &
   \cellcolor[HTML]{F3F3F3}DirectSAM~\cite{subobject}  \texttt{ZS} &
   \cellcolor[HTML]{F3F3F3}.712 &
   \cellcolor[HTML]{F3F3F3}.942 &
   \cellcolor[HTML]{F3F3F3}.449\\
 &
  \cellcolor[HTML]{F3F3F3}DirectSAM~\cite{subobject}  \texttt{FT} &
  \cellcolor[HTML]{F3F3F3}.925 &
  \cellcolor[HTML]{F3F3F3}.973 &
  \cellcolor[HTML]{F3F3F3}.773 \\
 &
  \cellcolor[HTML]{E2ECFD}DirectSAM-RS \texttt{ZS} &
  \cellcolor[HTML]{E2ECFD}.639 &
  \cellcolor[HTML]{E2ECFD}.803 &
  \cellcolor[HTML]{E2ECFD}.504 \\
\multirow{-6}{*}{\begin{tabular}[c]{@{}c@{}}SLSD~\cite{slsd}\\ (Coastline)\end{tabular}} &
  \cellcolor[HTML]{E2ECFD}DirectSAM-RS \texttt{FT} &
  \cellcolor[HTML]{E2ECFD}\textbf{.958} &
  \cellcolor[HTML]{E2ECFD}\textbf{1.000} &
  \cellcolor[HTML]{E2ECFD}\textbf{.895} \\ \bottomrule
\end{tabular}%
}
\end{table}

\begin{table}
\centering

\resizebox{\linewidth}{!}{%
\begin{tabular}{cc}
    \begin{minipage}{0.65\linewidth}
        \centering
        \caption{Performance comparison on LoveDA validation set.}
        \label{tab:pre-weights}
        \begin{tabular}{ccc}
            \toprule
            \textbf{\begin{tabular}[c]{@{}c@{}}Model\\ Architecture\end{tabular}} &
              \multicolumn{1}{c}{\textbf{\begin{tabular}[c]{@{}c@{}}Pretraining\\ Data\end{tabular}}} &
              \textbf{ODS} \\ \midrule
            LSTM-CNN~\cite{LSTMCNN} &      ILSVRC~\cite{ILSVRC}                           & .416 \\
            ConvLSTM~\cite{ConvLSTM} &    Pascal-VOC~\cite{pascalvoc}                            & .459 \\
            LAVT~\cite{lavt}     &        ImageNet22K~\cite{imagenet}                        & .595 \\
            RRSIS~\cite{rrsis}    &          ImageNet22K~\cite{imagenet}                      & .627 \\ \hline
                    \rule{0pt}{2ex}
                     & \multicolumn{1}{c}{Cityscapes~\cite{Cityscapes}} & .663 \\
            \multirow{-2}{*}{\begin{tabular}[c]{@{}c@{}}SegFormer~\cite{xie2021segformer} + \\ Prompter\end{tabular}} &
              \multicolumn{1}{c}{\cellcolor[HTML]{E2ECFD}\textbf{SA-1B~\cite{SAM}}} &
              \cellcolor[HTML]{E2ECFD}\textbf{.694} \\ \bottomrule
            \end{tabular}%
    \end{minipage}
    &
    \begin{minipage}{0.6\linewidth}
        \centering
        \caption{Ablation study on SA-1B pretrained weights.}
        \label{tab:pre_category}
            \begin{tabular}{c
            >{\columncolor[HTML]{E2ECFD}}c cc}
            \toprule
            \textbf{Class} & \textbf{\begin{tabular}[c]{@{}c@{}}w/\\ SA-1B\end{tabular}} & \textbf{\begin{tabular}[c]{@{}c@{}}w/o\\ SA-1B\end{tabular}} & $\Delta (\%)$ \\ \midrule
            Road        & .806 & .757 & {\color[HTML]{FE0000} -6.1}  \\
            \rule{0pt}{1.84ex}
            Agriculture & .668 & .625 & {\color[HTML]{FE0000} -6.4}  \\
            \rule{0pt}{1.84ex}
            Barren      & .666 & .606 & {\color[HTML]{FE0000} -9.0}  \\
            \rule{0pt}{1.84ex}
            Water       & .711 & .639 & {\color[HTML]{FE0000} -10.1} \\
            \rule{0pt}{1.84ex}
            Forest      & .756 & .673 & {\color[HTML]{FE0000} -11.0} \\
            \rule{0pt}{1.84ex}
            Building    & .759 & .654 & {\color[HTML]{FE0000} -13.8} \\ \bottomrule
            \end{tabular}%
    \end{minipage}
\end{tabular}%
}
\end{table}

Following HED\cite{HED}, we used the following evaluation metrics in the experiment: Optimal Dataset Scale (ODS) and Optimal Image Scale (OIS). Based on the $d_{max}$ index and image size $S$, the pixel tolerance $T$ is calculated by~$ T=EvenCeil(S\times d_{max})$. EvenCeil rounds the input $S \times d_{max}$ up to the nearest even integer. The ODS and OIS were calculated under the tolerance $T$ setting. Additionally, we employed Line Intersection over Union with a 3-pixel dilated kernel (LineIoU@3), following the evaluation metrics of BUBE. 








\begin{figure}[t]
\centering
\includegraphics[width=\columnwidth]{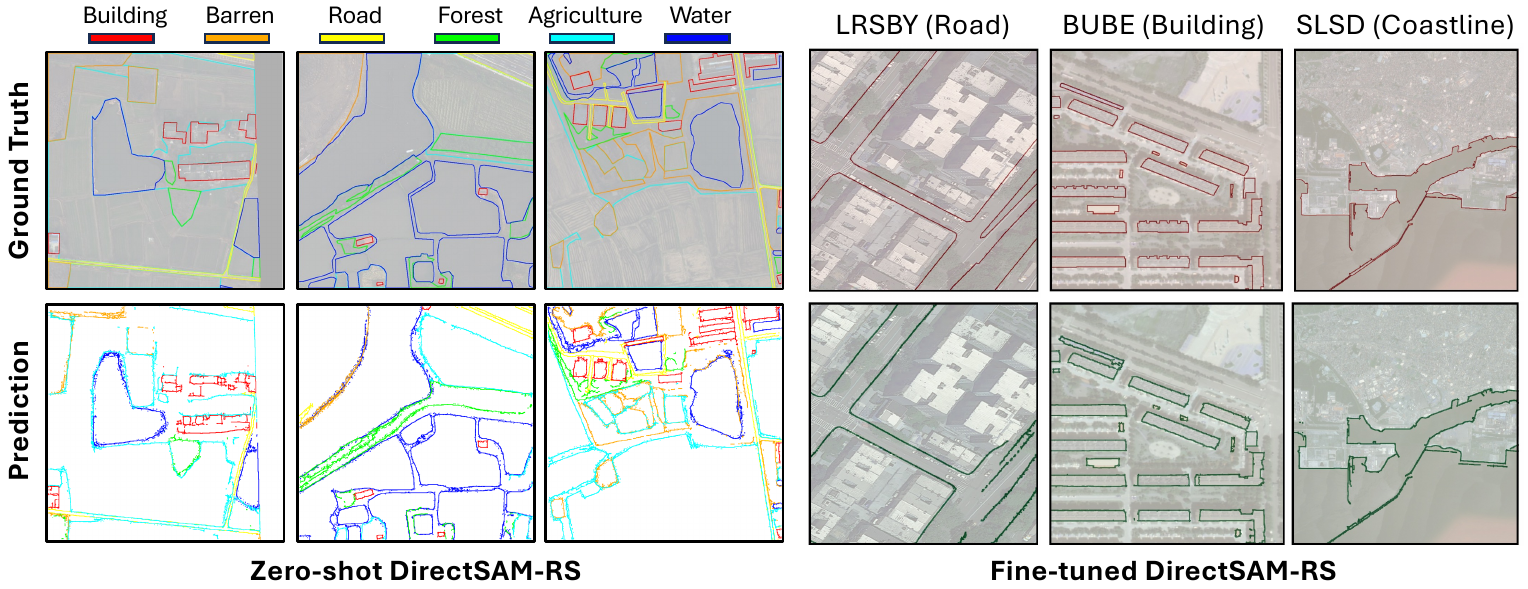}
\caption{\textbf{Inference examples of both zero-shot and fine-tuned DirectSAM-RS.} Zero-shot DirectSAM-RS (left) demonstrates its ability to flexibly adjust the semantic target according to the given prompt, while fine-tuned DirectSAM-RS (right) produces accurate contours for specific classes.}
\label{fig:vis}
\end{figure}

\begin{figure}[t]
\centering
\includegraphics[width=\columnwidth]{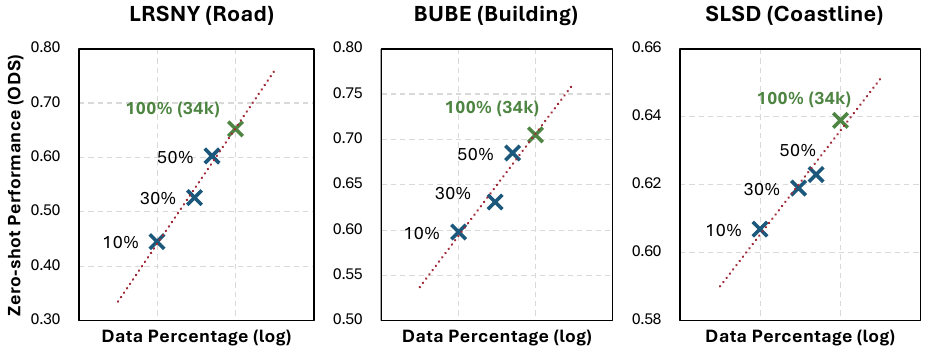}
\caption{\textbf{Data scaling experiment}. We used different-sized subsets of \texttt{RemoteContour-34k} to train the model and observed that model performance increases \textit{linearly} as the data scales exponentially.}
\label{fig:scale_data}
\end{figure}

\subsection{Benchmarking DirectSAM-RS}
We assessed the performance of DirectSAM-RS in both zero-shot and fine-tuning settings. As shown in Table~\ref{tab:main}, DirectSAM-RS achieved state-of-the-art performance across several downstream contour extraction benchmarks.

\paragraph{Zero-shot (\texttt{ZS}) setting}

We evaluated our DirectSAM-RS model in a zero-shot setting on three downstream contour extraction tasks, achieving strong performance without task-specific fine-tuning, as shown in Table~\ref{tab:main}. The improvement between DirectSAM-RS~\texttt{ZS} and DirectSAM~\texttt{ZS} highlighted the significant impact of integrating textual semantics, demonstrating that language information provides generalization potential for DirectSAM-RS. 

\paragraph{Fine-tuning (\texttt{FT}) setting}

Furthermore, we fine-tuned DirectSAM-RS with each benchmark dataset training split and evaluated it on the validation split. The results, presented in Table~\ref{tab:main}, revealed that DirectSAM-RS significantly surpasses previous road, coastline, and building extraction SOTA methods, achieving notable 21\%, 5\%, and 7\% improvement in ODS metrics respectively. The visualization inference result is shown in Figure~\ref{fig:vis}. The powerful performance on different practical landmark extractions showcased that DirectSAM-RS possesses strong application value.

\subsection{Ablation of DirectSAM SA-1B Pretraining} 
The pretraining weights ablation was conducted on the LoveDA validation set. As shown in Table~\ref{tab:pre-weights}, the DirectSAM-RS initialized from weights pretrained on SA-1B (DirectSAM-1800px-0424\footnote{\url{https://huggingface.co/chendelong/DirectSAM-1800px-0424}}) achieved the best ODS metric compared with existing methods pretrained on the traditional dataset. Therefore, we conducted SA-1B pretrained weights ablation study on each category of LoveDA validation set. Table~\ref{tab:pre_category} shows that DirectSAM-RS initialized from SA-1B weights outperforms the version without SA-1B weights by a large margin on every landmark category.

\subsection{Importance of Scaling-up Pretraining Data}

We conducted pretraining using three different fractions of the data from each category within each sub-dataset: 50\%, 30\%, and 10\%. This approach was used to quantify \texttt{RemoteContourRS-34k}'s impact on zero-shot contour extraction performance. Figure~\ref{fig:scale_data} illustrates that DirectSAM-RS demonstrates improvement when using 100\% of the pretraining data. However, there is still potential for further enhancement in zero-shot setting performance.




\section{Conclusion and Future Works}
\label{sec:majhead}


We introduce DirectSAM-RS for semantic contour extraction in remote sensing imagery. Its state-of-the-art performance can be attributed to two key factors: a large and diverse dataset with rich semantic information, and an attached prompter that enables the absorption of shared knowledge across different contour extraction tasks. Looking ahead, several promising directions for future work emerge. First, our experiments suggest that model performance has not yet been saturated with increasing data size. Expanding the dataset further could potentially yield additional improvements. Additionally, while we have demonstrated success in building, road, and coastline extraction, there are numerous other remote sensing tasks that could benefit from DirectSAM-RS.  Finally, the model's strong zero-shot performance suggests potential for few-shot learning in real-world scenarios with limited labeled data.

\bibliographystyle{IEEEtran}

\bibliography{strings,refs}

\end{document}